\documentclass{article}

\usepackage{arxiv}
\usepackage[utf8]{inputenc} 
\usepackage[T1]{fontenc}    
\usepackage{hyperref}       
\usepackage{url}            
\usepackage{booktabs}       
\usepackage{amsfonts}       
\usepackage{nicefrac}       
\usepackage{microtype}      
\usepackage{lipsum}
\usepackage{graphicx}
\usepackage{float}
\graphicspath{ {./images/} }

\title{Embedded Systems and Computer Vision Techniques utilized in Spray Painting Robots: A Review}

\author{
 Soham Shah \\
  Department of Information and Communication Technology\\
  Adani Institute of Infrastructure Engineering\\
  Ahmedabad, Gujarat, India - 382421 \\
  \texttt{sohamshah.ict18@aii.ac.in} \\
   \And
 Siddhi Vinayak Pandey \\
  Department of Electrical Engineering\\
  Adani Institute of Infrastructure Engineering\\
  Ahmedabad, Gujarat, India - 382421 \\
  \texttt{siddhipandey.ele17@aii.ac.in} \\
  \And
 Archit Sorathiya \\
 Department of Information and Communication Technology\\
  Adani Institute of Infrastructure Engineering\\
  Ahmedabad, Gujarat, India - 382421 \\
  \texttt{architsorathiya.ict18@aii.ac.in} \\
   \And
 Raj Sheth \\
 Department of Information and Communication Technology\\
  Adani Institute of Infrastructure Engineering\\
  Ahmedabad, Gujarat, India - 382421 \\
  \texttt{rajsheth.ict18@aii.ac.in} \\
   \And
 Alok Kumar Singh \\
 Department of Electrical Engineering\\
  Adani Institute of Infrastructure Engineering\\
  Ahmedabad, Gujarat, India - 382421 \\
  \texttt{alok.singh@aii.ac.in} \\
     \And
 Jignesh Thaker \\
 Department of Mechanical Engineering\\
  Adani Institute of Infrastructure Engineering\\
  Ahmedabad, Gujarat, India - 382421 \\
  \texttt{jignesh.thaker@aii.ac.in} \\
  
}

\begin{document}
\maketitle
\begin{abstract}
The advent of the era of machines has limited human interaction and this has in-creased their presence in the last decade. The requirement to increase the effectiveness, durability and reliability in the robots has also risen quite drastically too. Present paper covers the various embedded system and computer vision methodologies, techniques and innovations used in the field of spray-painting robots. There have been many advancements in the sphere of painting robots utilized for high rise buildings, wall painting, road marking paintings, etc. Review focuses on image processing, computational and computer vision techniques that can be applied in the product to increase efficiency of the performance drastically. Image analysis, filtering, enhancement, object detection, edge detection methods, path and localization methods and fine tuning of parameters are being discussed in depth to use while developing such products. Dynamic system design is being deliberated by using which results in reduction of human interaction, environment sustainability and better quality of work in detail. Embedded systems involving the micro-controllers, processors, communicating devices, sensors and actuators, software to use them; is being explained for end-to-end development and enhancement of accuracy and precision in Spray Painting Robots.  
\end{abstract}


\section{Introduction}
In the era of growing robotics, one needs to focus on the technological advancement by which the performance complexity of the robot is reduced and the feasibility of implementing the same increases. In today’s world, there is a huge requirement of automation in painting applications. Use spray painting robots for it, and for that the intensity of paint, accuracy and precision of the painted path, paint and energy consumption, cost efficiency and durability should be looked upon carefully before the development. Hence, a detailed study of these diverse topics is necessary for manufacturing products of similar application. This paper provides brief and sufficient insights about the latest trends and technologies currently in use or under research in the field of image processing and embedded computing and benefits of using spray painting techniques over manual methods specially in rare location. Various Image Processing algorithms, methods and techniques that are used in Spray Painting Robots are being discussed. The insights about sundry components of the embedded system including the sensors, actuators, micro-controllers, processors etc. are deliberated. All the technologies that are conferred are presented in adequate detail and in a lucid manner.

\section{Image Processing and Computer Vision}
\label{sec:headings}
Image Processing is a technique that is used to performing various operations on a digitized image and aimed to enhance the quality, manipulation of the image, identi-fying the objects or activities inside the image feed, analyzing and drawing out cer-tain results and assumptions out of it, etc. by using various mathematical models and transforms to be applied on the pixel-level matrix of the image.
Image-Processing techniques can be applied significantly in the embedded systems [1], resulting into precise and efficient outputs to the overall working of the system. The live video-feed is being processed and required operations are applied on the image field inside the processor; providing a virtual view for the control system and can perform desired tasks upon that; eradicating the control dependent only on the physical-parameters and static programming methods. Image Processing can be ef-fectively used for fully automated and dynamic Painting applications [2] by the high- pressure air spray nozzles; which get triggered when certain conditions or favorable parameters are detected in the video-feed; hence providing intelligence to the system.
\subsection{Object Detection}
Object detection is a Computer Vision [3] technique that deals with detecting instances of semantic objects of predefined classes for the applications of recognition of the target image and hence can take certain actions upon the static, dynamic or live image-feed into the system. An Object Detection detects the proba-bility of an object in an image depending on the previously trained model for the particular object to be detected. In Robotics application, in real time, apart from the detection of the object, shape determination, accurate tracking [4], object localization, reinforcement learning [6] can be used. 

\subsubsection{Deep Learning Methods}
One of the widely used transfer-learning algorithms for live image detection is YOLO [8]. “You Look Only Once (YOLO)” algorithm involves fetching all the pixels are taken all at the same time as an input to the deep learning model. YOLO V3[9] algorithm has around 80 different classes which also includes precise detection of person, bus, bicycle, traffic signs, car, motorbike etc. like obstacles that may come in the way of the automated spray-painting robot. It might be computationally expensive to detect many classes at the same time, so GPU-assisted processors must be used. Pretrained Models have gained popularity due to easy deployment, better architecture and high accuracy. Some other models can be used for Object Detection are Mobile Net SSD [10], R-CNN [11], Mask R-CNN [12], OverFeat, YOLOv1, etc. and for image classification InceptionV3 [13], Google Net [14], EasyNet, ResNet etc. can be used.
\par 
To identify the objects, the classical, yet widely used method is making use of Deep Learning [15]. To detect the particular set of objects in the image, the dataset of the cropped images of the target objects are used for training [16] the model. It is passed through different layers of the neural net [17]. The highest-probability class denotes the detected object. The weights of the neural network can be saved after the training for future usage. By this, the customized objects can be trained and can classify objects like potholes, uneven surfaces, traffic signs, road markings (where it has to paint) etc.

\subsection{Color Detection}
Color Detection was classically done by using the color-intensity sensors, which based on their intensity values could predict the color of the object; in which the sensor contain a white light emitter to illuminate the surface, and by the help of activation of the three filters (red, blue and green) color is classified. The Color detection sensors don’t work in the absence of light, require frequent calibration, has short range (<40cm) making them obsolete for today’s industrial applications. Computer Vision has proved to be more efficient solution for color detection [18]; which involves required mathematical operations on the image, binarization, gray-scaling, masking, thresholding, bitwise operations etc.
\par 
First, the image is acquired in a grayscale form and then stored in the system’s memory. To perform color detection on RGB image, we can perform in MATLAB [19] or OpenCV. HSV color space can be used over the RGB space because of the efficient separation of the image luminance from the color information. A threshold-ed image is generated which has only the HSV ranges of the upper and lower levels of the specific color, which is called Masking [20]. To obtain the desired color detected image, the threshold image is operated bitwise AND on the host image; and so hence only the common color ranges are found in the final image. The detected color in the image is highlighted by the bounding boxes and then this information it can be used for further I/O operations on the Robot. Once the system is properly calibrated, the camera along with detection of the color of the road to be painted (white, black, yellow etc.) but also object detection, edge capturing, contouring etc. tasks can be per-formed on the same system; which considerably supports dynamic systems if required processing power is made suffice.

\begin{figure} 
    \centering
    \includegraphics{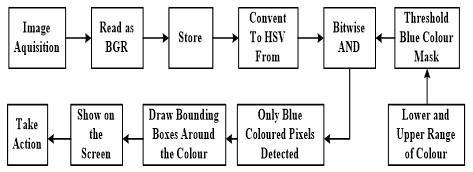}
    \caption{The figure shows the flow diagram of ‘Blue’ color detection.}
\end{figure}

\subsection{Edge Detection}
Edge Detection [21] is a mathematical model to find the edges around the target by detecting discontinuities in brightness of the images obtained. In Spray Painting system, there is a high possibility of luminance intensity variations and blurring of the frames in the live video feed. 
\par The Edges in an image are the significant local level changes, which occur on the boundary between two different regions; which is usually associated with a discontinuity in the image intensity or the first derivative of image intensity. These discontinuities are broadly of two types as explained in [22]; (1) step discontinuities; where the image intensity changes abruptly (2) line discontinuities; where the image intensities changes abruptly but then returns to the starting values within some finite distance. To detect the discontinuous significant local points in the image, discrete approximation to the gradient; which is the measure of the change in the function used for edge detection is applied on the pixel array of the target image. The samples of some continuous function of image intensity are stored in the array data structure. The Gradient is the two-dimensional equivalent of the first derivative, which is stored as a vector. The edge detection involves : 

\subsubsection{Image Filtering and Enhancement}
Image filtering [23] is a process in which the unwanted noises from the image are removed and overall visual color quality can be improved.  For instance, the captured image of the target is blur and to produce a clear image, Mean Filter is used; which performs average smoothening on the image by taking the average of each pixel value that surrounds it. There are Linear Smoothing filtering techniques like Box blur, Hann Window, Gaussian Blur etc. and Non-linear Smoothing techniques like Median filtering, Binary Morphological Operations, Min/Max Filters, etc. Also, some special techniques are used for smoothing like Spatial filters (can be applied on a dynamic system), Temporal filters etc. for noise reduction and quality improvisation. Image Enhancement [24] is used complimentary with filtering, which provides improvised sharpening and smoothening of the features. Enhancing of digital images leads to better object detection, segmentations, masking, color identifications and edge de-terminations. Some of the popular techniques are Histogram equalization, Adaptive Histogram equalization, Fuzzy Logic Technique, Nuro Fuzzy Technique, Unsharp Masking, Contrast Stretching, Log transformation, Local Enhancement etc. are used as per application [25].

\subsubsection{Detection and Localization}
The points with strong edge content are only considered as edges and they are localized in the image by using contours or bounding boxes. There are various detectors developed in the past two decades. The Edge detectors are of two types (1) first derivative operators – Sobel Operator, Roberts Operator, Prewitt Operator etc. (2) second derivative operators – Laplacian Operator, Gaussian Edge detector, Canny Edge Detector, etc. It is evaluated on the factors like probability of false edges, probability of missing edges and error in estimation of edge angle. Localization includes the formation of the Contours [26] around the detected pixels of the edges and hence giving the dynamic system the intelligence to focus only on the region inside the contour; which here in our case can be fully automating the painting by detecting the edges on the roads, pavements etc. and painting only in the region inside the contour. Painting inside the contour area is yet a challenging task; for which a bright red-color light can be emitted from the robot using VL618OX-LIDAR distance sensor’ to the area to be painting. The moment sharp red color is detected inside the contour region by the system [27,28]. By this, the system is made dynamic and can take decisions on basis the situational analysis where we do not need to explicitly program, efficient painting and avoiding all the menace caused due to the programmed delay time for the mo-tors to operate while painting [29].

\begin{figure} 
    \centering
    \includegraphics{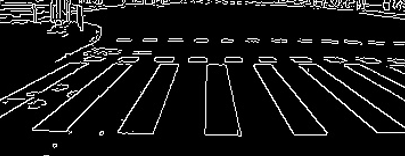}
    \caption{The above figure demonstrates the edge detection and contour formation of an example image of zebra crossing on road [30]}
    \end{figure}

\section{Embedded System}
\subsection{Software}
The Spray-Painting Robot as a whole system requires frequent decision making, in-formation processing, data storage, communication, I/O operations, data tracking and many such operations all at the same time [31]. To handle these kinds of operations, programming software are used to structure and organize the work of the whole system. The Software are used for programming the controllers, processors, performing image processing, training models, making conditional and logical decisions in real time, communicating wirelessly via Bluetooth, Radio frequency methods etc.  By software programming, the resource utilization is made efficient [32,33].
\par 
Software that are used for the Spray-Painting Robot depends upon the functionalities for which it is developed. For microcontrollers, Arduino, Atmel Studio, Codeblocks, etc. For implementation of computer vision algorithms into the Robotic system; Python, C++, MATLAB etc. [34] languages are used. These high-level programming languages have wide range of libraries for image processing like OpenCV, Scikit-image, SciPy, PIL, Pycairo, Pgmagick, SimpleITK, Mahotas etc. and has inbuilt function for various image processing tasks. To train and deploy deep learning models, neural networks and adaptive learning methods, the libraries and high-end APIs such as TensorFlow, Keras, PyTorch, Numpy, Pandas etc. can be used. For the communication between hardware and software, ROS is one of the popular alternatives available for integrating the system. ROS (Robot Operating System) [35] is a framework or a Meta OS which can also be used for communication and as an intermediary between the sensors and actuators; where the algorithms, system calls, drivers etc. are handled, which is operated on an OS (Linux) platform. ROS can be used for integrating all the hardware components of the Robot and controlling by one single platform. Moreover, PLC/HMI [36], can be used to satisfy the time demand requirements and matches industrial standards and hence multiple industrial components can be incorporated with the same system for hybrid applications.
\subsection{Electronics}
\subsubsection{Microcontrollers}
For small scale and explicit programmable systems, Microcontrollers serve as the most affordable as well as the fastest control system for the sensors, actuators, communicators on the Robot. Arduino family [37] controllers, ATMEGA family, etc. microcontrollers are used which has no memory or processor allocated for the execution of the commands [38]. Implementation of Image Processing Algorithms has to be done by external camera feed device and cannot be integrating with the control system of the Robot in this case.
\subsubsection{Microprocessors}
The applications that involve high computations, precise decision, computational ability, memory storage and access and frequent read-write operations, implementation of Image Processing and Computer Vision techniques etc. the microprocessors are used as the central control unit of the Robot. They provide high accuracy and durability in the application and thus leading to better decision-making for the Robot. Some of the microprocessors that can be incorporated are Raspberry PI family [39], NVidea JETSON NANO etc. for performing real time decision making and information processing [40].
\subsubsection{Sensors}
The sensors can be used in various domains for Spray Painting Robot. Environmental-sensor Modules [41] that include Temperature Sensors (LM35 etc.), Smoke detector sensors (MQ2 etc.), Relative Humidity Sensors (DHT11 etc.), etc. can be used for safety measures of the Robot and interrupt the painting process if any harmful signal values are received by the Environmental Sensor Modules [42]. The Ultrasonic sensors can be placed on the side of the robot facing the ground, which can detect the potholes [43], sinkholes etc. on the roads; hence ensuring safety for the robot from damage. Line Tracking Sensors [44] can also be incorporated to detect lines on the Roads and hence path can be well maintained. For some less precise and intrinsic applications, color sensors like white-line detectors etc. can also be used for Color Detection, which can be connected on the bottom of the Robot body. For carrying out image processing, Cameras of high quality must be mounted on the Robot. Use of Tachometer can be done to measure speed of the robot [45,46].
\subsubsection{Communicating Devices}
For short range applications, Bluetooth devices prove to be the most efficient as they can be easily connected with the mobile devices or computers. HC-05 is an example of such Bluetooth device family as in [46,47]. BLE (Bluetooth Low Energy) devices have gained popularity due to considerably low power consumption. For long range, HC-01 can be used which has 1KM line of sight range. NodeMCU or ESP8266 Wi-Fi modules can also be used for communication for comparatively longer range than Bluetooth devices. For long range communications, XBee or LORA can be used, which works on Radio frequency protocols.
\subsubsection{Industrial Tools}
PLC or Programmable Logic Controller is an industrial grade controller which can be used for industrially compliant results of the spray painting[48,49]. It has fast processing and precision in performance to enhance results. It can handle a variety of peripheral devices, including various sensors and camera. HMI or Human Machine Interface is a device which can display real time information and can be used to alter the state of the machine whenever required and thus the name Human Machine Interface [50]. In spray painting robots it can be used to change the on/off timer or speed, or any other such parameters as required by simply touch or hardware-based input system. The HMI is connected to the controller and sends commands to it. 
\begin{figure}[H]
\centering
    \includegraphics{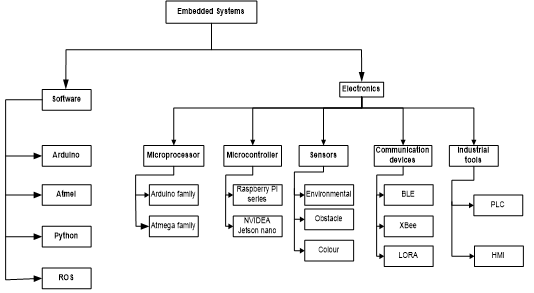}
    \caption{The above figure summarizes the embedded systems for spray painting robots}
    \end{figure}
    
\section {Conclusion}
The human interaction has been replaced by automation drastically and thus the demand for increasing the efficiency, durability and reliability in Robotics applications is substantial. When developing and designing Spray painting robots, many aspects come into consideration relating to the electrical and electronics system, mechanical and dynamics and computation and software technologies. This paper produces a detailed description about the recent advancements and research in areas which includes information about various image processing techniques, pretrained models, image analysis and dynamic and real time decisions by the system, which can be used as a reference significantly for manufacturing Spray-Painting Robots and similar applications in order to increase the accuracy and effectiveness to a greater extent. A thorough analysis of the various software tools, electronics actuators, sensors, microcontrollers and microprocessors cumulated to design and develop the Embedded Systems for Spray Painting Robots is being presented and also provides insights for justified selection as per requirement of the application. Further-more, the present review accumulates the different technologies and parameters which result in reduction of cost, human interaction, environment sustainability and quality of work in detail. Thus, by integrating the various technologies for development of automated or semi-automatic dynamic systems using the latest technologies; resourceful, durable and efficient robots can be developed. 
\section{References}
1.	Mashal, Vaishali.: Design and Implementation of Image Processing Applications in Em-bedded System. International Journal of Innovative Research in Computer and Communica-tion Engineering 4(8), 15556–15562 (2016).
\par 2.	Kohilan, M Joseph. et al.: Computer Vision Controlled Automated Paint Sprayer Using Image Processing and Embedded System. International Journal of Innovative Research in Science, Engineering and Technology 3(2), 156–160 (2014).
\par 3.	Wu, Juan. et al: Research on Computer Vision-Based Object Detection and Classification. International Federation for Information Processing 1, 183–188 (2013).
\par 4.	Tiwari, Mukesh. et al: A Review of Detection and Tracking of Object from Image and Vid-eo Sequences. International Journal of Computational Intelligence Research 13, 745–765 (2017).
\par 5.	Alizadh, Shima.: Convolutional Neural Networks for Facial Expression Recognition. Jour-nal 2(5), 99–110 (2016).
\par 6.	Caicedo, J. C., and Lazebnik, S: Active object localization with deep reinforcement learning In: Proceedings of the IEEE International Conference on Computer Vision pp. 2488–2496. IEEE, New York City (2007).
\par 7.	S, Geethapriya. et al.: Real-Time Object Detection with Yolo. International Journal of Engi-neering and Advanced Technology 8(3), 578–581 (2019).
\par 8.	Redmon, Joseph. et al: You Only Look Once: Unified, Real-Time Object Detection. arXiv, 1–110 (2016).
\par 9.	Howard, G. Andrew. et al: MobileNets: Efficient Convolutional Neural Networks for Mo-bile Vision Applications. arXiv, 1–9 (2017).
\par 10.	Girshick, Ross. et al: Region-based Convolutional Networks for Accurate Object Detection and Segmentation. IEEE 2(5), 1–16 (2016).
\par 11.	Szegedy, Christian. et al: Rethinking the inception architecture for computer vision . arXiv, 1–10 (2015).
\par 12.	He, Kaiming. et al: Mask R-CNN. In: 2017 IEEE International Conference on Computer Vision, pp. 2980–2988. IEEE, New York City (2017).
\par 13.	Szegedy, Christian . et al: Going deeper with convolutions. arXiv 1–12 (2014).
\par 14.	Zheng, Peng. et al: Object Detection with Deep Learning: A Review. arXiv 1–21 (2019).
\par 15.	Szegedy, Christian. et al: Deep Neural Networks for Object Detection. SemanticScholar, 1–9 (2016).
\par 16.	Ding, Sheng. et al: Research on Daily Objects Detection Based on Deep Neural Network. In: IOP Conference Series: Materials Science and Engineering 2018, vol. 322, pp. 1–6. Elsevier, Netherlands (2018). 
\par 17.	Ding, Sheng. et al: Research on Daily Objects Detection Based on Deep Neural Network: IOP Conference Series: Materials Science and Engineering 2018, vol. 322, pp. 1–6. Else-vier, Netherlands (2018). 
\par 18.	Goel, Vishesh. et al: Specific Color Detection in Images using RGB Modelling in MATLAB. International Journal of Computer Applications  161, 38–42 (2017).
\par 19.	Prabhakar, Arshi. et al.: DIFFERENT COLOR DETECTION IN AN RGB IMAGE. In-ternational Journal of Development Research 7(8), 14503–14506 (2017).
\par 20.	Habiur, Rahman. et al: Segmentation of color image using adaptive thresholding and mask-ing with watershed algorithm. Segmentation of color image using adaptive thresholding and masking with watershed algorithm (2013).
\par 21.	R, Muthukrishnan.: EDGE DETECTION TECHNIQUES FOR IMAGE SEGMENTATION. International Journal of Computer Science and Information Technology (IJCSIT) 3(6), 259–267 (2011).
\par 22.	Jain, Ramesh., Kasturi, Rangachar. et al: Edge Detection Machine Vision 1995, LNCS, vol. 9999, pp. 140–185. McGraw-Hill, Delhi (1995).
\par 23.	Gupta. et al: Image Filtering Algorithms and Techniques: A Review. International Journal of Advanced Research in Computer Science and Software Engineering 3(10), 198–202 (2013).
\par \par 24.	Potnis, Anjali. et al: A Review on Image Enhancement Techniques. International Journal of Engineering and Applied Computer Science 2(7), 232–235 (2017).
\par 25.	Chandwadkar, Radhika.: Comparison of Edge Detection Techniques.  2(5), 99–110 (2016).
\par 26.	Chandwadkar, Radhika. et al: Comparison of Edge Detection Techniques. In First Interna-tional Conference on artificial intelligence and cognitive: Bapi, Raju Surampudi, (eds.) CONFERENCE 2019, pp 99-110 Springer, Heidelberg (2019). 
\par 27.	Ehsan, Javanmardi. et al: Autonomous Vehicle self-localization based on abstract map and multi-channel LiDAR in Urban Area. IATSS Research 43(1), 1-13 (2018).
\par 28.	Kazuhiko, Kibayashi: Analysis and prevention of traffic fatalities and injuries from the per-spective of forensic medicine. IATSS Research 43(2), 69-70 (2019).
\par 29.	Liang, Wang; Yihuan, Zhang; Jun, Wang: Map based localization method for autonomous vehicles using 3d – LIDAR. IFAC - PaperOnLine 50(1), 276-281 (2017).
\par 30.	ZEBRACROSSING https://www.collinsdictionary.com/images/full/zebracrossin/onlinedictionary, Last Accessed 2020.04.24 .  
\par 31.	Naresh, Vurokonda; B. Thirumala, Rao:  A Study on Storage Security Issues in Cloud Computing. Procedia Computer Science 92, 128-135 (2016). 
\par 32.	Nidhi; Km Archana, Patel:  An Efficient and Scalable Density Based Clustering Algorithm for Normalize Data. Procedia Computer Science 92, 136-141 (2016). 
\par 33.	Youguo, Li; Haiyan, Wu: A clustering Method Based on K-Map Method. Procedia Com-puter Science 25, 1104-1109 (2012). 
\par 34.	COLLINSDICTIONARY https://www.collinsdictionary.com/images/full/zebracrossing.jpg?version=4.0.42 /images, Last Accessed 2020.04.25 . 
\par 35.	Author, F., Author, S.: Title of a proceedings paper. In: Editor, F., Editor, S. (eds.) CONFERENCE 2016, LNCS, vol. 9999, pp. 1–13. Springer, Heidelberg (2016). 
\par 36.	Kajan, Martin.: Control of Automated Guided Vehicle with PLC SIMATIC ET200S CPU. American Journal of Mechanical Engineering 1(7), 343–348 (2013).
\par 37.	Louis, Leo.: WORKING PRINCIPLE OF ARDUINO AND USING IT AS A TOOL FOR STUDY AND RESEARCH. International Journal of Control, Automation 1(2), 21–29 (2016).
\par 38.	MICROCONTROLLER https://www.tutorialspoint.com/microprocessor/microcontrollersoverview.htm 
\par /Microcontroller Overview, Last Accessed 2020.04.27 . 
\par 39.	Shilpashree, K.S. et al: Implementation of Image Processing on Raspberry Pi. International Journal of Advanced Research in Computer and Communication Engineering 4(5), 199–202 (2015).
\par 40.	INFORMATIONS https://www.tutorialspoint.com/microprocessor/microcontrollersoverview.htm 
\par /Microcontroller, Last Accessed 2020.04.28 . 
\par 41.	Alhmiedat, Tareq. et al: A Survey on Environmental Monitoring Systems using Wireless Sensor Networks. JOURNAL OF NETWORKS 10(11), 606–615 (2015).
\par 42.	IOTSENS https://iotsens.com/sensors/environmental-sensor/ /Environmental Sensor, Last Accessed 2020.04.29 . 
\par 43.	Madli, Rajeshwari. et al: Automatic Detection and Notification of Potholes and Humps on Roads to Aid Drivers. IEEE SENSORS JOURNAL 15(8), 4313–4318 (2015).
\par 44.	Sharma, Deepak. et al: Line Tracking Robotic Vehicle. International Journal of Science and Research (IJSR), 551–553 (2015).
\par 45.	SENSIRION https://www.sensirion.com/en/environmental-sensors/ / Environmental is-sues, Last Accessed 2020.04.29 . 
\par 46.	OMRON https://www.components.omron.com/solutions/mems-sensors/environment-sensor / E - Sensors, Last Accessed 2020.04.30 . 
\par 47.	Singh, Pratibha. et al: A Modern Study of Bluetooth Wireless Technology. International Journal of Computer Science, Engineering and Information Technology (IJCSEIT) 1(3), 55–63 (2011).
\par 48.	Cotta, Anisha. et al: WIRELESS COMMUNICATION USING HC-05 BLUETOOTH MODULE INTERFACED WITH ARDUINO. International Journal of Science, Engineer-ing and Technology Research (IJSETR) 5(4), 869–872 (2016).
\par 49.	GUIDE https://www.globalspec.com/learnmore/building 
\par /industrial
\par 50.	Reddy, Harshvardhan. et al.: PLC based Robot Manipulator Control using Position based and Image based Algorithm. Global Journal of Researches in Engineering: H Robotics and Nano -Tech 17(1), (2017).

\end{document}